\documentclass{bmvc2k}
\usepackage{comment}
\usepackage{graphicx}
\usepackage{caption}
\usepackage{adjustbox}
\usepackage[noblocks]{authblk}
\usepackage{ifsym}
\usepackage{float}
\title{Multimodal Semi-Supervised Learning for 3D Objects}

\addauthor{Zhimin Chen}{zhiminc@clemson.edu}{1}
\addauthor{Longlong Jing}{ljing@gradcenter.cuny.edu}{2}
\addauthor{Yang Liang }{lyang1@ccny.cuny.edu}{2}
\addauthor{YingLi Tian }{ytian@ccny.cuny.edu}{2}
\addauthor{Bing Li (Corresponding author) }{bli4@clemson.edu}{1}

\addinstitution{
Clemson University\\
 Clemson, USA
}
\addinstitution{
The City University of New York,\\
 New York, USA
}

\runninghead{ZHIMIN ET AL.}{Multimodal Semi-Supervised Learning for 3D Objects}

\begin{document}
\begin{titlepage}

\maketitle

\begin{abstract}
In recent years, semi-supervised learning has been widely explored and shows excellent data efficiency for 2D data. There is an emerging need to improve data efficiency for 3D tasks due to the scarcity of labeled 3D data. This paper explores how the coherence of different modalities of 3D data (e.g. point cloud, image, and mesh) can be used to improve data efficiency for both 3D classification and retrieval tasks. We propose a novel multimodal semi-supervised learning framework by introducing instance-level consistency constraint and  a novel multimodal contrastive prototype (M2CP) loss. The instance-level consistency enforces the network to generate consistent representations for multimodal data of the same object regardless of its modality. The M2CP maintains a multimodal prototype for each class and learns features with small intra-class variations by minimizing the feature distance of each object to its prototype while maximizing the distance to the others. Our proposed framework significantly outperforms all the state-of-the-art counterparts for both classification and retrieval tasks by a large margin on the modelNet10 and ModelNet40 datasets.
\end{abstract}

\section{Introduction}
Due to the scarcity of large-scale labeled dataset, in recent years, the semi-supervised learning method has been drawing wide attention and showing the great potential of boosting up the performance of networks by jointly training on both limited labeled data and a large number of unlabeled samples~\cite{lee2013pseudo, miyato2018virtual, radosavovic2018data, tarvainen2017mean,  zhai2019s4l}. It can significantly improve the data efficiency of the neural network training by leveraging unlabeled data, such as   Pseudo-Labeling~\cite{lee2013pseudo} which uses the confident prediction of the network as labels to further optimize the network, and FixMatch~\cite{sohn2020fixmatch} which optimizes the network to predict consistency output for images with different augmentation from the same image.

Although many semi-supervised learning methods have been proposed for 2D-related image recognition tasks, semi-supervised learning for 3D-related tasks has not been widely explored. Moreover, we observe that methods that were originally proposed for 2D-related tasks are not able to achieve comparable performance for 3D tasks (e.g., 3D object classification). Different from image and video data, 3D data usually consists of different modalities. The multimodal coherent among these modalities contain rich semantic information of the objects, which can be  utilized to advance the 3D semi-supervised learning. Motivated by multimodal learning in other fields~\cite{jing2020cross, simonyan2014two, chen2020simple}, we propose a multimodal semi-supervised learning framework based on two novel constraints including instance-level consistency and multimodal contrastive prototype (M2CP) constraints.
\begin{figure}[t!]
\centering
\begin{minipage}[t]{0.4\textwidth}
\centering
\captionsetup{width=.99\linewidth}
\includegraphics[scale=0.1]{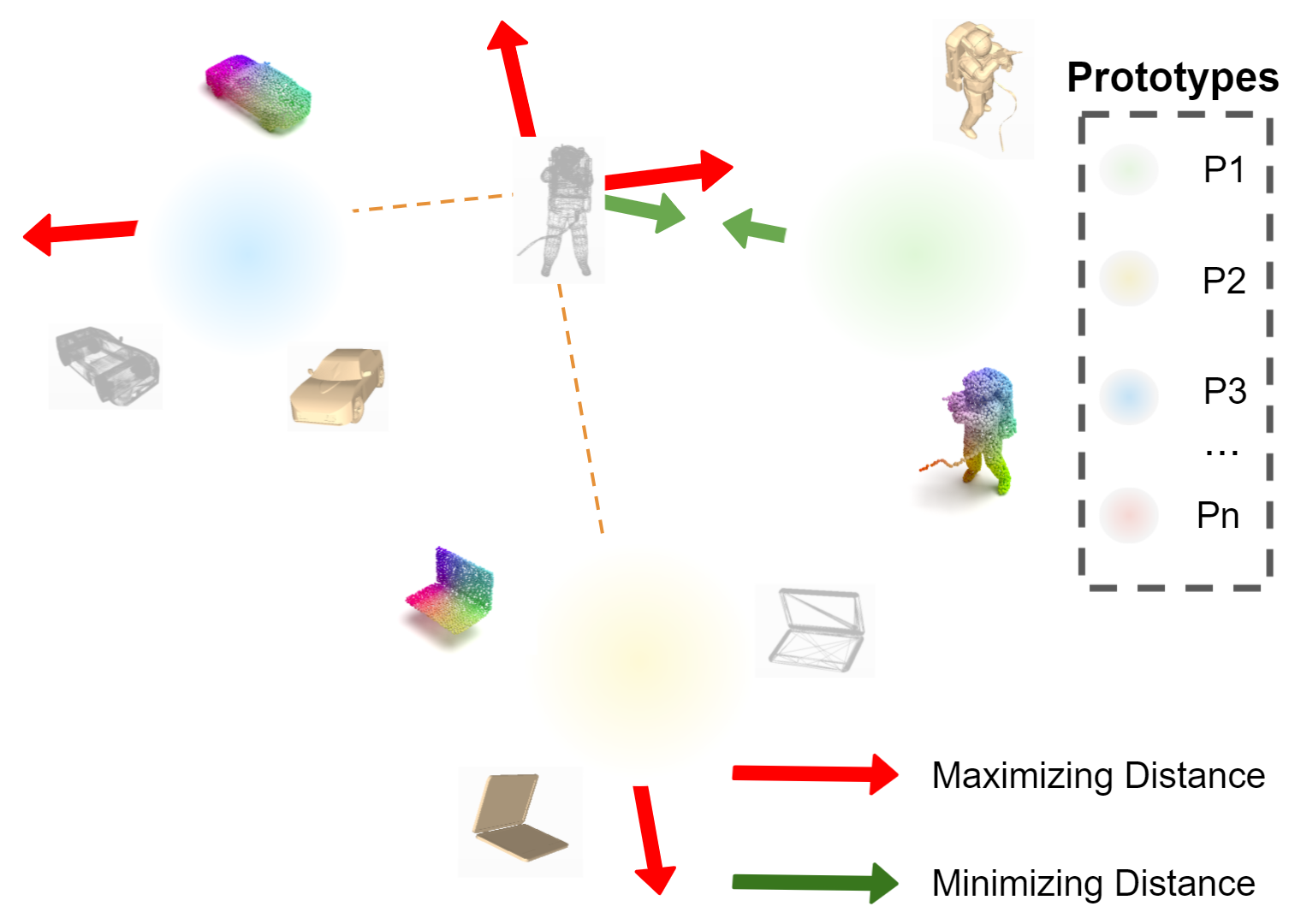}
\vspace*{1mm}

\caption{The M2CP maintains a multimodal prototype for each class and learns features with small intra-class variations by minimizing the feature distance of each object to its prototype while maximizing the distance to the others.}

\label{fig:loss}
\end{minipage}
\begin{minipage}[t]{0.45\textwidth}
\centering
\captionsetup{width=.85\linewidth}

\includegraphics[scale=0.27]{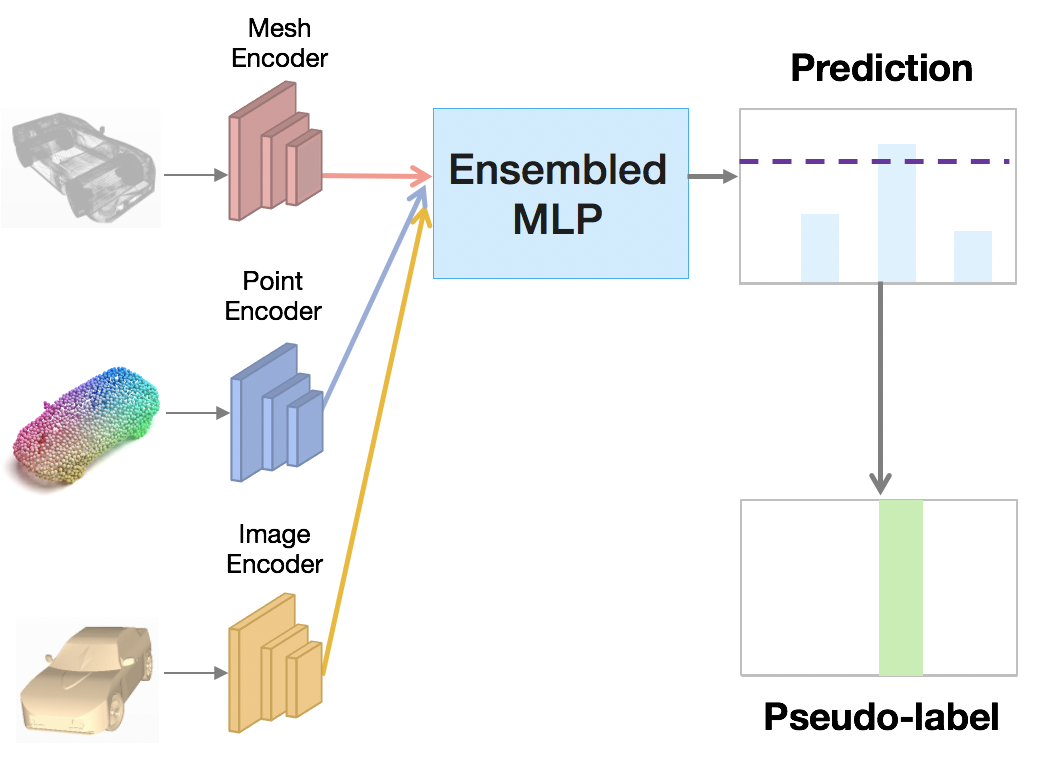}
\vspace*{5mm}

\caption{For the unlabeled samples, the features from multiple modalities are aggregated together by an Ensembled-MLP to produce more reliable and consistent pseudo-labels which will be used by the our proposed M2CP loss.}

\label{fig:psd}
\end{minipage}
\end{figure}

As shown in Fig.~\ref{fig:loss}, the M2CP maintains a multimodal prototype for each class and learns to minimize the feature distance of each object to its prototype while maximizing the distance to the others. By minimizing the M2CP loss, the features from different classes are more separable while the features from the same class are closer, and it can potentially benefit the classification and retrieval tasks. Extensive experiments are conducted on two public benchmark datasets (i.e. ModelNet10 and ModelNet40) for both classification and retrieval tasks with three different modalities including point cloud, mesh, and image. Our key contributions are summarized as follows: 1) A novel multimodal contrastive prototype (M2CP) loss is proposed to learn the coherent embedding across multi-modalities. It can simultaneously minimize the intra-class distances and maximize the inter-class distances of embedding features by utilizing prototypes in semi-supervised learning. 2) We propose a novel multimodal semi-supervised framework that further encompasses instance-level consistency loss which enforces the network to generate consistent predictions for multimodal data of the same object regardless of its modality. 3) Our comprehensive experiments and ablation studies demonstrated that our proposed method significantly outperforms the state-of-the-art semi-supervised learning methods for both 3D object classification and retrieval tasks across point cloud, mesh, and image modalities on the ModelNet10 and ModelNet40 datasets.

\section{Related Work}

\textbf{3D Object Classification:} 3D object classification is a fundamental task for 3D understanding~\cite{qi2017pointnet, qi2017pointnet++,  thomas2019kpconv, wu2019pointconv, xu2018spidercnn, zhao2019pointweb, zhao2020point, xu2020weakly,xu2020weakly}. Since 3D objects usually can be represented in different modalities while each one has its advantages, many methods have been proposed for different modalities including image~\cite{su2015multi}, mesh~\cite{feng2019meshnet, hanocka2019meshcnn, Schult_2020_CVPR, wiersma2020cnns}, point cloud~\cite{li2019deepgcns, qi2017pointnet, qi2017pointnet++, dgcnn}, etc. However, these methods normally only focus on one modality. Our method is specifically  designed to explore the coherence of different modalities of 3D data for semi-supervised learning.

\textbf{Semi-Supervised Learning:} Many semi-supervised learning methods have been proposed, but most of them focused on image-related tasks and some of them are very difficult to be directly transferred to other tasks or data~\cite{berthelot2019mixmatch, chen2020big, ghosh2021data, lee2013pseudo, miyato2018virtual, tarvainen2017mean, oliver2018realistic}. These methods usually learn by enforcing networks to produce consistent predictions for different views of the same data~\cite{ berthelot2019remixmatch, kuo2020featmatch, sohn2020fixmatch}, by minimizing the entropy of the predictions on unlabeled data~\cite{lee2013pseudo}, or by other types of regularization~\cite{zhai2019s4l}. With the rapid development of self-supervised learning, various self-supervised tasks are used as auxiliary loss for semi-supervised learning~\cite{cai2021exponential, si2020adversarial,  zhai2019s4l}.  Recently, the 3D semi-supervised learning for tasks including 3D object classification~\cite{sanghi2020info3d, song2020semi}, 3D semantic segmentation~\cite{chen2021semi, mei2019semantic,ouali2020semi} and 3D object detection ~\cite{zhao2020sess, wang20213dioumatch} start to draw attention from the community. Most of these methods mainly use the data from one modality and the coherence of multimodal data is normally ignored. By utilizing the multimodal data with our proposed constraints, we set a comprehensive benchmark for both 3D classification and retrieval tasks and our proposed model significantly outperforms the most recent state-of-the-art methods~\cite{song2020semi, sanghi2020info3d}.

 \textbf{Self-Supervised 3D Feature Learning:}  Due to the advantage of utilizing unlabeled data, more and more 3D self-supervised learning methods have been proposed~\cite{xie2020pointcontrast, hou2021pri3d, hou2021exploring, sanghi2020info3d, yang2018foldingnet, zhao20193d, sharma2020self}. The existing self-supervised learning methods normally learn features by accomplishing pre-defined tasks such as contrasting~\cite{jing2021self}, context prediction~\cite{hassani2019unsupervised}, orientation prediction ~\cite{poursaeed2020self}, etc. The self-supervised learning methods do not require any labels, therefore, they can be used as an auxiliary task for other tasks to help the network to learn more representative features.

\textbf{Multimodal Feature Learning:} Multi-modality has been widely studied in many tasks. Different modalities capture data from different perspectives while the features from multiple modalities are usually complementary with each other. A typical example is the two-stream network~\cite{feichtenhofer2016convolutional, simonyan2014two} for video action classification task which fuses feature extracted from RGB video clips and features from optical flow stacks~\cite{dosovitskiy2015flownet, ilg2017flownet}. The similar idea has been explored in many other tasks including RGBD semantic segmentation~\cite{qi20173d}, video object detection~\cite{zhu2017flow}, 3D object detection, and sentiment analysis~\cite{hazarika2020misa} etc. In this paper, we propose a novel multimodal semi-supervised learning framework with two constraints to jointly learn from both labeled and unlabeled data for 3D classification and retrieval tasks.
\vspace*{-5mm}



\section{Multimodal Semi-Supervised Learning}

 Fig.~\ref{fig:framework} is the overview of our framework, with labeled and unlabeled three modalities of point cloud, mesh, and image as input. The framework consists of three components including supervised learning on labeled data, instance-level consistency learning on unlabeled data, and multimodal contrastive prototype (M2CP) learning on both labeled and unlabeled data. The generalized formulation of our proposed model is described in the following sections. 

\begin{figure*}[tb]
\centering
\includegraphics[width = 0.95\textwidth]{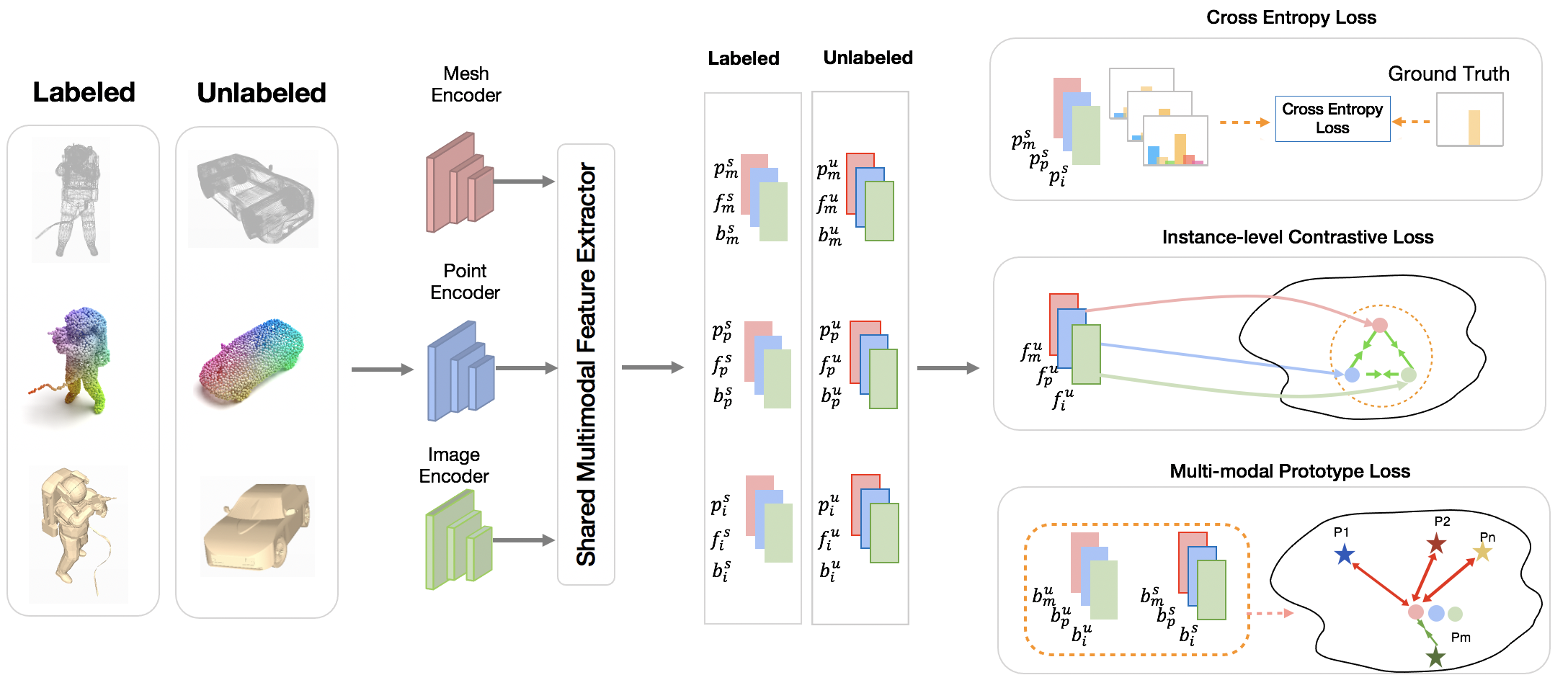}
\vspace*{3mm}
\setlength{\belowcaptionskip}{-10pt}

\caption{An overview of the proposed framework for multimodal semi-supervised learning. Our model is jointly trained on data with three different modalities from both labeled and unlabeled data. Three loss functions are employed to train the network including (1) a regular cross-entropy loss on the labeled data to learn discriminative features, (2) an instance-level consistency loss to enforce network to predict consistency features for multimodal data, and (3) our proposed novel multimodal contrastive prototype loss which minimize the intra-class distances and maximize the inter-class distances of multimodal features simultaneously.} 

\label{fig:framework}
\end{figure*}

\subsection{Problem Setup}

Given a set of limited labeled data $X_L$ and a large amount of unlabeled data $X_U$, the labeled data $X_L$ is formulated as:\begin{equation}
    X_L=\left\{t_i^l\right\}_{i=1}^{N_l}, t_i^l =\left(s_i^l, y_i\right), s_i^l=\left\{x_i^{lm}\right\}_{m=1}^M,
\end{equation} where $t_i^l$ is a instance of labeled data containing $M$ modalities $\left\{x_i^{lm}\right\}_{m=1}^M$ with label $y_i$.

The unlabeled data $X_U$ is formulated as:\begin{equation}
    X_U=\left\{t_i^u\right\}_{i=1}^{N_u}, t_i^u =\left(s_i^u\right) , s_i^u=\left\{x_i^{um}\right\}_{m=1}^M,
\end{equation} where each instance in the unlabeled data $t_i^u$ only contains data $\left\{x_i^{um}\right\}_{m=1}^M$ in $M$ different modalities. Our model is trained on both $X_L$ and $X_U$ for multimodal semi-supervised learning with the proposed constraints.

\subsection{Representation Learning}

Given each data sample $\textbf{x}^{m}_i$ from modality $m$, a network $F_m$ that designed for modality $m$ produces a general hidden feature vector as: 
\begin{equation}
    H^{m}_i = F_m(x^{m}_i),
\end{equation} while $H^{m}_i$ is the general hidden feature to represent the data $x^{m}_i$. There are $m$ different feature encoders in total and one for each modality.


To implicitly enforce the consistency, two shared multimodal feature encoders are employed across all the modalities to map each generally hidden feature vector $H^{m}_i$ into two outputs. One is a task-specific feature vector and the other is the classification prediction as:
\begin{equation}
    E^{m}_i = W(H^{m}_i), ~~~~ {\hat{y}}^{m}_i = G(H^{m}_i),
\end{equation} where $E^{m}_i$ is a task-specific feature vector to represent the data $x^{m}_i$ which will be used for instance-level consistency learning and ${\hat{y}}^{m}_i$ is the corresponding classification prediction based on $x^{m}_i$. By sharing $W()$ and $G()$ across all the modalities, the network implicitly enforces the feature extractors $F_1, ..., F_m$ to generate consistent features for different modalities of the same object.

Therefore, for each data instance $t_i$, no matter it is from labeled dataset $X_L$ or unlabeled data $X_U$, task-specific feature vectors {$E^{1}_i$, ... $E^{m}_i$}, and predictions {${\hat{y}}^{1}_i$, ..., ${\hat{y}}^{m}_i$} are obtained with the shared multimodal encoder from a set of general hidden feature vectors {$H^{1}_i$, ... $H^{m}_i$}. Our proposed constraints are optimized over these extracted features and predictions.
\subsubsection{Supervised Training}

The supervised training over the labeled data $X_L$ helps the network to learn discriminative features. For each sample $x_i$ from the labeled set $X_U$, the cross-entropy loss is calculated between all the predictions ${\hat{y}}^{m}_i$ and the fused prediction result $y^{f}$. The corresponding ground truth label:
\begin{equation}
    L_{e}=-\frac{1}{N}(\sum_{i=1}^{N}(\sum_{m=1}^{M}y_i\cdot log({\hat{y}}^{m}_i) + y_i\cdot log({y}^{f}_i)), 
\end{equation} in which the cross-entropy loss from $M$ and different modalities fused prediction  are averaged over $N$ samples to optimize the network.

\subsubsection{Instance-Level Consistency Learning}

Normally there are two ways to perform the instance-level consistency learning either from prediction level or from feature level. The existing methods mainly optimize networks by enforcing consistency in the prediction level such as enforcing networks to make the same classification prediction over two views of the same data sample. However, the predictions of networks are often very noisy especially when the labeled data is very limited, and the consistency learning from the prediction level inevitably involves severe noises during the training. Relying on the multimodal attributes of 3D data, we propose to regularize the network with instance-level consistency learning from the feature level in which the network is trained to predict consistent features $E_m$ for data from different modalities of the same object.

We propose to directly maximize the similarity of multimodal features from the same object while minimizing the similarity of multimodal features from different objects. Given the recent progress of contrastive learning, the contrastive loss is employed for the instance-level consistency learning over the task-specific features $E_m$ extracted by our shared multimodal feature encoder. Given any two modalities $x_i^{m1}$ and $x_i^{m2}$ from a total of $M$ different modalities of the data $x_i$, the multimodal features are firstly extracted and represented as $E^{m1}_i$, ... $E^{m2}_i$ from 
the shared multimodal feature encoder. Then the instance-level cross-model consistency loss is optimized through:
{
\small
\begin{equation}
\label{eq:instance}
\hspace{-4mm}
\mathcal{L}_{i}(E^{m1}_i, E^{m2}_i)\!\!= \!-\!\!\log\!\frac{h\big(E^{m1}_i, \!E^{m2}_i\big)}{\sum\limits_{\substack{k=1}}^{B}\!\!\!h\big(E^{m1}_i,\!E^{m2}_k\!\big)},
\end{equation} } 
where $h(\mathbf{u},\mathbf{v}) = \exp\big(\frac{\mathbf{u}^\top \mathbf{v}}{||\mathbf{u}||_2||\mathbf{v||_2}}/\tau\big)$ is the exponential of cosine similarity measure, $\tau$ is the temperature, and  $B$ is the batch size. Each mini-batch consists of data from multiple modalities, and the instance-level consistency loss is calculated on combinations of any of the two modalities. 

\subsection{Multimodal Contrastive Prototype Learning}


The instance-level consistency focuses on the consistency of the instance-level feature representation, but the relations among instance and classes are not utilized. To thoroughly utilize the information hidden on categories, we propose the multimodal contrastive prototype loss to jointly constraint the intra-class and the inter-class variations by using both labeled and unlabeled data.

A prototype $P_i$ is defined for each class $i$ and is learned and updated through the training. Each prototype represents the semantic center for one class in the feature space. For each data sample $x_i^m$, the distance of features of $x_i^m$ to its corresponding prototype $P_i$ is minimized while the distance with the rest of the prototypes is simultaneously maximized. Formally, for data sample data $x_i^{m}$ with its general hidden features $H^{m}_i$ and prototype $P_i$, the multimodal contrastive prototype loss is formulated as:
{
\small
\begin{equation}
\hspace{-3mm}\mathcal{L}_{p}= \!-\!\log\!\frac{g(H^{m}_i, \!P_i)}{\sum\limits_{\substack{k=1}}^{C}\!\!g(H^{m}_i, \!P_k)}, 
\end{equation}
}
\normalsize while $g(H^{m}_i, \!P_i) = exp(-\frac{\| H^{m}_i - P_i \|_2^2}{\tau})$, $C$ is the number of prototypes ,and $\tau$ is the temperature.

For each data sample from the labeled data, its category is required for the M2CP loss to know the distance with which prototypes should be minimized. To fully utilize a large amount of unlabeled data, we extend this loss to both labeled and unlabeled data by generating pseudo-labels for the unlabeled data during the training. A simple idea would be to select the confident predictions (i.e. if the prediction $\hat{y^{m}_i}$ is larger than a threshold) and then use these confident predictions as labels for these selected unlabeled data as data to jointly train the multimodal contrastive prototype loss. However, this may lead to inconsistent pseudo-labels for the different modalities of the same object.

To generate consistent pseudo-labels for different modalities of the same object, we propose to fuse the general features $F_m$ from different modalities to obtain an object-level prediction which will then be used as pseudo-labels for all the modalities of this object, as shown in Fig.~\ref{fig:psd}. Formally, the features {$H^{1}_i$, ..., $H^{m}_i$} from $m$ modalities are aggregated concatenated together to represent the object $t_i$ to get the final prediction:
\begin{equation}
    y^{f}_i = K(H^{1}_i, ..., H^{m}_i),
\end{equation} while $K$ has  two MLP layers network to predict $y_i$ based on the concatenated features of multiple modalities. Having access to multiple modalities, the prediction $y_i$ is more reliable compared to the predictions from a single modality. The pseudo-labels are further created as:
\begin{equation}
    \hat{y}^c_i = \begin{cases} 1, \quad  \text{if}~max(y^{f}_i) \ge \delta \\ 
    0,~~\text{otherwise} \end{cases},
\end{equation} 
while $\delta$ is the pre-defined threshold. By generating pseudo-labels for unlabeled data, our proposed multimodal contrastive prototype loss is able to be trained by both labeled data $X_L$ and large-scale unlabeled data $X_U$.

Our entire framework is jointly optimized on both labeled and unlabeled data with the three weighted loss functions as:\begin{align}\label{eq:loss_combined}
Loss = \alpha L_{e} +\beta L_{i} + \lambda L_{p},
\end{align}
\noindent
where $\alpha$, $\beta$, and $\lambda$ are the weights for each loss term.


\subsection{Architecture}

\textbf{Feature Extractor:} There are three feature extractors and one for each modality. Due to the powerful ability to learn image features, ResNet~\cite{ResNet} is used as backbone networks for image feature extractors. For the point cloud feature extractor, the DGCNN \cite{dgcnn} is employed due to its powerful ability to capture local structures with the KNN graph module from the point cloud. The MeshNet~\cite{feng2019meshnet} is chosen as a feature encoder for mesh modality, and it takes the $n$ faces and its normal vectors as inputs. The shared multimodal feature encoder consists of two parallel branches while one has three MLP layers with a size of $512, 256, 256$ and the other with a size of $512, 256, C$ while $C$ is the number of classes.
\vspace*{-3mm}

\section{Experimental Results}

\textbf{Datasets:} The proposed framework is evaluated on two datasets including ModelNet10 \cite{wu20153d} and ModelNet40~\cite{wu20153d} with different percentages of labeled samples. The ModelNet40 dataset is a 3D object benchmark that consists of $12,311$ samples belong to $40$ different categories while $9,843$ for training and $2,468$ for testing. The ModelNet10 dataset consists of $4,900$ samples belong to $10$ categories with $3,991$ for training and $909$ for testing.

\textbf{Setup and Training Details}
On the ModelNet40 dataset, our model is trained with an SGD optimizer with a learning rate of $0.01$ for a total of $10,0000$ iterations. The learning rate is reduced by $90\%$ every $4,0000$ iterations. On the ModelNet10 dataset, the model is trained for a total of $6,0000$ iterations and the learning rate starts from $0.01$ and is reduced by $90$\% every $2,0000$ iterations.  The weights are $1$, $2$, and $9$ for cross-entropy, instance-level contrastive loss, and multimodal contrastive prototype loss respectively. For all the experiments, a batch size of $48$ is used while half of them are labeled data and the other half are unlabeled data.

\subsection{Performance on Semi-Supervised 3D Object Classification}
Since there are only few existing semi-supervised learning methods specifically designed for the 3D object classification task and without a comprehensive semi-supervised benchmark, we first compare with methods that were originally designed for 2D semi-supervised learning but apply them to the 3D object classification task. We compare with the supervised baseline and three different semi-supervised methods include Pseudo-Labeling (PL)~\cite{lee2013pseudo}, FixMatch~\cite{sohn2020fixmatch}, and S4L~\cite{zhai2019s4l} under the same settings. Worth to note that the FixMatch-based methods achieve state-of-the-art performance on many semi-supervised tasks \cite{zhao2020sess}, \cite{sohn2020simple}, \cite{xiong2021multiview}, \cite{xie2020humble}. For a fair comparison, all the methods are using the same backbones and data augmentations as our methods, and only one modality is available during the inference phase for all the methods.

The performance comparison with the above-mentioned four methods on the ModelNet40 and ModelNet10 datasets for 3D object classification are shown in Table~\ref{tab:Classification}. To extensively evaluate the performance, we compare with the other methods under different percentages ($2$\%, $5$\%, and $10$\%) of labeled data for both datasets. The following conclusions can be drawn from the comparison: 1) Since the state-of-the-art semi-supervised learning methods were mainly designed for image-based tasks, they perform well on the image modality while having a negligible impact on the point cloud and mesh modalities. Directly adapting these methods to the 3D classification task obtains unsatisfied results due to the lack of specific constraints. 2) By jointly training our framework with multiple modalities, the performance with all three modalities is significantly improved regardless of the amount of labeled data. 3) With the different amounts of labeled data, our method consistently significantly improves the classification performance and outperforms all three state-of-the-art semi-supervised learning methods. These results confirm the advantage of our proposed multimodal semi-supervised method.

\begin{table*}[ht]
\adjustbox{width=350pt,height=800pt,keepaspectratio=true}
{

	\begin{tabular}{c|c|c|c|c|c||c|c|c|c|c|c}
	    \hline
    	\multicolumn{6}{c||}{ModelNet40} & \multicolumn{6}{c}{ModelNet10}  \\
    	\hline
    	Test  & Baseline & PL & FixMatch  &S4L &Ours&     	Test  & Baseline & PL & FixMatch  &S4L &Ours\\
        Modality& & \cite{lee2013pseudo}&\cite{sohn2020fixmatch}& \cite{zhai2019s4l}& & Modality& &\cite{lee2013pseudo} &\cite{sohn2020fixmatch}& \cite{zhai2019s4l}& \\

    	\hline
    	\multicolumn{12}{c}{2\% of Labeled data} \\
    	\hline
    	Image    &69.61  &72.20  &75.69  &80.96 &\textbf{82.78} &
    	Image    &73.57  &74.04 &70.44  &80.51  &\textbf{85.46} \\
    	Point    &63.21  &65.68  &65.56  &68.64  &\textbf{79.86}&
    	Point    &77.31  &78.41  &78.74  &81.61  &\textbf{84.36}\\
    	Mesh     &48.18  &49.33  &52.59  &51.72  &\textbf{78.81}&
    	Mesh     &66.41  &63.00  &61.44  &65.40  &\textbf{86.13}\\

    	\hline
    	\multicolumn{12}{c}{5\% of Labeled data} \\
    	\hline
    	Image    &82.29  &83.18  &85.49  &84.12  &\textbf{88.61}&
    	Image    &83.70  &86.56 &84.36  & 88.33 &\textbf{92.14}\\
    	Point    &76.62  &79.74  &79.38  &78.69 &\textbf{85.29}&
    	Point    &83.59  &86.34  &85.13 &84.91  &\textbf{89.87}\\
    	Mesh     &71.19  &73.49  &72.76  &77.02 &\textbf{86.51}&
    	Mesh     &80.51  &80.73  &82.60  &79.74 &\textbf{90.75}\\

        \hline
        \multicolumn{12}{c}{10\% of Labeled data} \\
        
        \hline
        Image  &   85.90&   86.95&   89.02&  87.16 & \textbf{91.61}&
        Image    &91.85  & 91.74  & 90.86 & 92.07 &   \textbf{93.95}\\
    	Point  &   82.86&   84.04&   84.81&  83.75 & \textbf{88.49}&
    	Point    &87.44   &88.00    & 88.33  &88.16  & \textbf{91.63}\\
    	Mesh   & 80.39   &  82.47  & 81.36  &  82.42 & \textbf{88.29}& Mesh     &84.25    &87.11    & 83.29  &81.83  &  \textbf{92.84}  \\
        \hline
	\end{tabular}
}

\vspace*{3mm}
 \caption{Performance comparison for the 3D object classification task with the state-of-the-art semi-supervised learning methods on the ModelNet40 and ModelNet 10 dataset with different percentages of labeled data.}
\label{tab:Classification}
\end{table*}
\vspace*{-5mm}

\begin{table*}[ht]
\adjustbox{width=350pt,height=800pt,keepaspectratio=true}{
	\begin{tabular}{c|c|c|c|c|c||c|c|c|c|c|c}
		\hline
    	\multicolumn{6}{c||}{ModelNet40} & \multicolumn{6}{c}{ModelNet10}  \\
    	\hline
    	Test & Baseline & PL & FixMatch  &S4L &Ours&     	Test & Baseline & PL & FixMatch  &S4L &Ours\\
        Modality&& \cite{lee2013pseudo}&\cite{sohn2020fixmatch}& \cite{zhai2019s4l}&& Modality&&\cite{lee2013pseudo} &\cite{sohn2020fixmatch}& \cite{zhai2019s4l}& \\
    	\hline
    	\multicolumn{12}{c}{2\% of Labeled data} \\
    	\hline
    	Image    &63.01  &62.84 &73.27  &73.13  &\textbf{{81.50}}&
    	Image    &68.90  &65.87 &73.34  &78.52  &\textbf{{83.82}}\\
    	Point    &55.43  &60.42  &61.57  &59.41 &\textbf{{78.45}}&
    	Point    &72.81  &76.42  &76.96 & 68.94  &\textbf{{84.85}}\\
    	Mesh     &50.50  &52.39  &53.37  &54.81  &{\textbf{80.31}}&
    	Mesh     &64.17  &70.99  &72.62  &67.23  &{\textbf{84.09}}\\
    	\hline
    	\multicolumn{12}{c}{5\% of Labeled data} \\
    	\hline
    	Image    &73.68  &74.38  &78.87  &74.90  &{\textbf{85.71}}&
    	Image    &79.00  &77.73  &82.49  &80.25  &{\textbf{87.67}}\\
    	Point    &57.92  &64.02  &63.09  &61.11  &{\textbf{82.05}}&
    	Point    &72.98  &76.60  &75.25 &69.39  &{\textbf{87.62}}\\
    	Mesh     &56.98  &63.94  &63.59  &58.81  &{\textbf{84.84}}&
    	Mesh     &75.81  &81.62  &79.00 &72.95 &{\textbf{88.43}}\\
        \hline
    	\multicolumn{12}{c}{10\% of Labeled data} \\
        \hline
    	Image    &78.15  &79.11 &82.24  &79.28  &{\textbf{86.96}}&
        Image    &85.33    &85.28    &87.68   &83.39  &{\textbf{90.74}}\\
    	Point    &60.20    &64.50    & 63.60  &64.96  & {\textbf{84.16}}  & 
    	Point    &72.93    &76.33    & 74.70 &71.18 &{\textbf{89.75 }}\\
    	Mesh     &60.20    &73.22   & 73.74  &72.02  &  {\textbf{84.29}}& 
    	Mesh     &81.01    &84.70    &78.03   &70.92 &{\textbf{90.61}}\\
        \hline
	\end{tabular}
}
\vspace*{3mm}
\setlength{\belowcaptionskip}{-30pt}

\caption{Performance comparison for the {3D object retrieval} task with other state-of-the-art semi-supervised learning methods on the ModelNet40 and ModelNet10 dataset with different percentages of labeled data.}
\label{tab:Retrieval}
\end{table*}

\vspace*{5mm}

\subsection{Performance on Semi-Supervised Object Retrieval}

We further evaluate the performance of the 3D object retrieval task and compare it with the state-of-the-art semi-supervised methods on both ModelNet40 and ModelNet10 datasets. Following the convention, the mean Average Precision (mAP) is used to indicate the performance.

We report the retrieval performance with different amounts of labeled data for all three modalities. As shown in Table~\ref{tab:Retrieval}, all these three state-of-the-art semi-supervised learning methods can only improve the performance on the image modality while the performances for other modalities sometimes are even worse than the baseline, which is probably due to the noises during the training. Benefited from our novel constraints, our method significantly improves the performance consistently for all the modalities by using different percentages of labeled data. These results demonstrate the effectiveness of our proposed framework and the generalizability in the 3D object retrieval task.

\subsection{Ablation Study} 

\begin{table}
\begin{minipage}[t]{0.5\linewidth}
	\begin{center}
	    \adjustbox{width=160pt,height=200pt,keepaspectratio=true}{
		\begin{tabular}{c|c|c|c|c}
		\hline
		{Modality} & $L_e$ & $L_e,L_i$ & $L_e, L_{p}$ & $L_e, L_i, L_{p}$\\
			\hline
            \multicolumn{5}{c}{3D Object Retrieval} \\
			\hline
			Image   & 78.15    &    76.74       &  85.26            &      {\textbf{86.96}} \\
			Point   & 60.20    &    70.40       &      82.51        &      {\textbf{84.16}} \\
			Mesh    & 60.20&74.91 & 78.04 &    {\textbf{84.29}}  \\
            \hline
			\multicolumn{5}{c}{3D Object Classification}     \\
			\hline
			Image    & 85.90 &89.34 & 89.79  &{\textbf{91.61}}  \\
			Point    & 82.86 &85.78 & 87.03&   {\textbf{88.49}}\\
			Mesh    & 80.39 & \textbf{88.61} & 85.82 & {88.29}\\
            \hline
		\end{tabular}
		}
	\end{center}
	\captionsetup{width=.9\linewidth}
	\caption{Ablation study for the combination of losses to the 3D object classification and retrieval tasks on ModelNet40 dataset with $10$\% of labeled data. }
		\label{tab:loss-ablation}
\end{minipage}
  \begin{minipage}[t]{0.5\linewidth}
   
   \centering

\begin{center}
\adjustbox{width=180pt,height=230pt,keepaspectratio=true}{
		\begin{tabular}{c|c|c|c|c}
		\hline
		{Modality} & Mesh-Image & Image-Point & Point-Mesh &All\\
			\hline
            \multicolumn{5}{c}{3D Object Retrieval} \\
			\hline
			Image   & 86.15  & 86.23  & ---    &{\textbf{86.96}} \\
			Point   & ---   & 82.90  &82.47   &{\textbf{84.16}} \\
			Mesh    & 81.27  & ---   & 79.68   &{\textbf{84.29}}  \\
            \hline
			\multicolumn{5}{c}{3D Object Classification}     \\
			\hline
			Image    & 89.95  & 90.48   & ---    &{\textbf{91.61}}  \\
			Point    & ---   & 87.48   & 86.47  &{\textbf{88.49}}\\
			Mesh    & 86.83   & ---   & 84.93    &{\textbf{88.29}}\\
            \hline
		\end{tabular}
		}
	\end{center}
	\captionsetup{width=.9\linewidth}

	\caption{Ablation study for the number of modalities to the 3D object classification and retrieval tasks on ModelNet40 dataset with $10$\% of labeled data. }
		\label{tab:modality-num-ablation}

\end{minipage}
\end{table}

\vspace*{-1mm}

\textbf{Ablation Study for Loss Functions:} Our proposed framework is jointly trained with three loss functions including cross-entropy loss $L_e$, instance-level consistency loss $L_i$, and a novel multimodal prototype loss $L_p$. To understand the impact of each loss term, we conduct ablation studies with four combinations of different loss functions including: 1) $L_e$, 2) $L_e+L_i$, 3) $L_e+L_p$, 4) $L_e+L_i+L_p$. We report the performance using different amounts of labeled data for both 3D classification and 3D retrieval tasks on the ModelNet40 dataset in Table~\ref{tab:loss-ablation}. From the result of Table~\ref{tab:loss-ablation}, we can draw the conclusion that: 1) When only the cross-entropy loss $L_e$ is used, the performance for both classification and retrieval tasks with all the modalities are very low due to the very limited labeled samples. 2) When the M2CP loss $L_p$ is jointly used with the cross-entropy loss $L_e$, the performances for all the tasks are significantly improved compared to the baseline, as well as significantly outperforms the performance of $L_e + L_i$. The results are consistent with our hypothesis since $L_i$ does not use the category information and only enforces the instance-level consistency, while the M2CP utilizes the category information to regularize the hidden features. 3) The best performances are achieved for all the tasks when all the three losses are used indicating that they are indeed complementary with each other.

\textbf{Ablation Study for Number of Modalities:} Compared to other semi-supervised methods, our method is designed to explicitly leverage the multimodal coherence of multimodal data with the proposed novel constraints. Our model is jointly trained from three Modalities including point cloud, mesh, and image. When more modality data are available, better performance should be achieved since more multimodal constraints are available to the networks. To verify the impact of the number of Modalities, we conduct experiments by training with three different modality combinations including (1) point cloud and image, (2) point cloud and mesh, and (3) image and mesh. The performance for both classification and retrieval tasks on the ModelNet40 dataset is shown in Table~\ref{tab:modality-num-ablation}. The performances for both tasks are the best when all three Modalities are used in training. This confirms our assumption that more Modalities can provide more constraints which produces better performance.
\vspace*{-5mm}

\section{Comparison with the State-of-the-Art 3D Semi-Supervised Methods}
\vspace*{-2mm}

To further demonstrate the capability of our proposed methods, we compare with the state-of-the-art methods~\cite{song2020semi, sanghi2020info3d} that are specifically designed for the 3D semi-supervised object classification task. The performance comparison with these methods on the ModelNet40 dataset is shown in Table ~\ref{tab:cotraining_comparison}. 

Our method significantly outperforms the state-of-the-art methods~\cite{song2020semi, sanghi2020info3d} with different modalities under different settings demonstrating the effectiveness of our proposed method. Our method outperforms the Info3D~\cite{sanghi2020info3d} by almost 9\% when only 2\% labeled data is available during training. The Deep Co-training~\cite{song2020semi} is specifically designed for 3D semi-supervised learning which mainly uses the consistency from the \textbf{prediction level} of the multimodal data as constraints however, the results are comparable with the 2D based method FixMatch~\cite{sohn2020fixmatch}. Relying on our proposed constraints learning directly from \textbf{feature level}, our model significantly outperforms state-of-the-art 2D and 3D semi-supervised methods by a large margin with both modalities under the same setting on the ModelNet40 dataset. Moreover, with our proposed novel M2CP loss directly optimizing from the feature level, our model is able to learn features with small intra-class variations which can achieve state-of-the-art results for the 3D retrieval tasks.

\vspace*{-5mm}

\begin{table}
\centering
	\begin{center}
\adjustbox{width=200pt,height=2400pt,keepaspectratio=true}{	   
		\begin{tabular}{|c|c|c|c|}
		    \hline
		    Method & Percentage & Modality &Accuracy  \\
			\hline
			Info3D~\cite{sanghi2020info3d}      & 2$\%$  &Point cloud & 71.06$\%$ \\
			\textbf{Ours}      & 2$\%$ &Point cloud &
			{\textbf{79.86$\%$}} \\
			\hline
			Info3D~\cite{sanghi2020info3d}      & 5$\%$  &Point cloud & 80.48$\%$ \\	
			\textbf{Ours}      & 5$\%$  &Point cloud & {\textbf{85.29$\%$}} \\
			\hline			Deep Co-training~\cite{song2020semi}      & 10$\%$    &Point cloud& 83.50$\%$ \\
            FixMatch~\cite{sohn2020fixmatch}      & 10$\%$    &Point cloud& 84.81$\%$ \\
			\textbf{Ours}      & 10$\%$  &Point cloud &{\textbf{88.49$\%$}} \\ 
			\hline
			Deep Co-training~\cite{song2020semi}     & 10$\%$    &Image & 89.00$\%$ \\
			FixMatch~\cite{sohn2020fixmatch}     & 10$\%$    &Image & 89.02$\%$ \\
			\textbf{Ours}    & 10$\%$    &Image& {\textbf{91.61$\%$}} \\
            \hline
		\end{tabular}}
		
	\end{center}
	\setlength{\belowcaptionskip}{-10pt}

	\caption{Comparison with the most recent state-of-the-art 2D and 3D semi-supervised methods~\cite{sanghi2020info3d, song2020semi, sohn2020fixmatch}. Our model significantly outperforms all of them with different modalities and different settings on ModelNet40.}
		\label{tab:cotraining_comparison}
\end{table}

\section{Conclusion}
\vspace*{-2mm}
We have proposed a novel multimodal semi-supervised learning method for 3D objects based on the coherence of multimodal data. The network jointly learns from both labeled and unlabeled data mainly using the proposed instance-level consistency and multimodal contrastive prototype (M2CP) constraints. Our proposed method remarkably outperforms the state-of-the-art semi-supervised learning methods and the baseline on both 3D classification and retrieval tasks. These results demonstrate that it is a promising direction to study how to apply the multimodal for 3D semi-supervised learning tasks. 
\vspace*{-5mm}

\section{Acknowledgement}

This work was supported in part by U.S. DOT UTC grant 69A3551747117.

\bibliography{egbib}
\end{titlepage}

\end{document}


\begin{titlepage}

\maketitle

This document contains the supplementary materials for "Multimodal Semi-Supervised Learning for 3D Objects".

\section{Comparison with Cross-Modal Center Loss}

\begin{table*}[!ht]
\setlength{\tabcolsep}{0.8mm}{
\begin{center}
\begin{tabular}{c|c|c|c|c|c|c}
\hline
Amount & \multicolumn{2}{c|}{2\%}  & \multicolumn{2}{c|}{5\%} & \multicolumn{2}{c}{10\%} \\ 
\hline
Loss  & $X_c$ & $X_p$ & $X_c$ & $X_p$ & $X_c$ & $X_p$ \\ \hline
\multicolumn{7}{c}{3D Cross-Modal Retrieval}  \\ \hline
Image2Point & $76.10$ &$\textbf{78.46}$ & $80.98$    & $\textbf{83.30}$   & $82.03$  & $\textbf{85.93}$ \\ 
\hline
Image2Mesh    & $76.47$    & $\textbf{80.05}$            & $82.47$ & $\textbf{85.04}$ & $83.57$ & $\textbf{87.56}$  \\ 
\hline
Point2Image   & \multicolumn{1}{c|}{78.03}                    & \multicolumn{1}{c|}{\textbf{79.28}}            & \multicolumn{1}{c|}{80.95}                    & \multicolumn{1}{c|}{\textbf{83.18}}            & \multicolumn{1}{c|}{82.69}                    & \multicolumn{1}{c}{\textbf{85.51}}            \\ \hline
Point2Mesh    &77.65 & \textbf{78.62}  &79.75   & \textbf{82.58} &81.23  &\textbf{85.00}\\ 
\hline
Mesh2Image    & \multicolumn{1}{c|}{77.30}    &\textbf{79.10}          & \multicolumn{1}{c|}{82.03}              & \textbf{84.73}           & \multicolumn{1}{c|}{84.00}             &\textbf{87.10}  \\ 
\hline
Mesh2Point  &75.20    & \textbf{77.42}  & 80.52   &\textbf{82.27}    &82.22 & \textbf{85.23} \\ \hline
\multicolumn{7}{c}{3D In-Domain Retrieval}
\\ \hline
Image & $78.84$ &$\textbf{81.50}$ & $83.23$    & $\textbf{85.71}$   & $84.48$  & $\textbf{86.96}$ \\ 
\hline
Point    & $76.08$    & $\textbf{78.45}$            & $79.67$ & $\textbf{82.05}$ & $83.08$ & $\textbf{84.16}$  \\ 
\hline
Mesh   & \multicolumn{1}{c|}{77.16}                    & \multicolumn{1}{c|}{\textbf{80.03}}            & \multicolumn{1}{c|}{82.43}                    & \multicolumn{1}{c|}{\textbf{84.84}}            & \multicolumn{1}{c|}{82.04}                    & \multicolumn{1}{c}{\textbf{84.29}}            \\ \hline
\multicolumn{7}{c}{3D Object Classification}                                    
\\ \hline
Image & $80.84$ &$\textbf{82.78}$ & $86.23$    & $\textbf{88.61}$   & $90.48$  & $\textbf{91.61}$ \\ 
\hline
Point    & $78.08$    & $\textbf{79.86}$            & $83.67$ & $\textbf{85.29}$ & $87.08$ & $\textbf{88.49}$  \\ 
\hline
Mesh   & \multicolumn{1}{c|}{77.16}                    & \multicolumn{1}{c|}{\textbf{78.81}}            & \multicolumn{1}{c|}{85.43}                    & \multicolumn{1}{c|}{\textbf{86.51}}            & \multicolumn{1}{c|}{86.04}                    & \multicolumn{1}{c}{\textbf{88.29}}  \\ \hline

\end{tabular}
\end{center}
\caption{Performance comparison of our proposed M2CP ($X_p$) loss with CMC ($X_c$) loss on the ModelNet40 dataset. Our proposed M2CP loss consistently outperforms the CMC loss in all three tasks.}
\label{tab:M2CP}}
\end{table*}


The Cross-Modal Center~\cite{jing2020cross} (CMC) loss is specifically proposed for cross-modal retrieval based on Center Loss~\cite{wen2016discriminative} which directly minimizes the intra-class distances of features from multiple modalities in the universal feature space. The CMC loss only minimizes the feature distance of each object to its prototype. Without any constraints about the relation of samples from different classes, the CMC loss may learn features with low inter-class variation. On the contrary, our proposed M2CP loss jointly minimizes the feature distance of each object to its prototype and maximizes the distance to the others. Therefore, M2CP can learn features with both low intra-class variation and high inter-class variation which further produces better classification and retrieval results.

To thoroughly compare the performance of these two losses, we conducted two sets of experiments, one is with M2CP  ($X_p$) loss and the other is with CMC loss ($X_c$), for the semi-supervised learning on the ModelNet40 dataset. We compared the performance on three different tasks including 3D cross-modal retrieval, 3D in-domain retrieval, and 3D object classification task. As shown in Table~\ref{tab:M2CP}, the proposed M2CP outperforms CMC loss for all these three tasks with three type percentages of labeled data. For some retrieval tasks such as Image2Point and Image2Mesh, our proposed M2CP outperforms CMC by more than $3$\% demonstrating its effectiveness.

\bibliography{egbib}
\end{titlepage}